\documentclass[conference]{IEEEtran}
\IEEEoverridecommandlockouts
\usepackage{cite}
\usepackage{amsmath,amssymb,amsfonts}
\usepackage{algorithmic}
\usepackage{graphicx}
\usepackage{amsthm}
\usepackage{textcomp}
\usepackage{xcolor}
\usepackage{subfigure}
\usepackage{tabularx}
\usepackage{booktabs}
\usepackage{longtable} 
\usepackage{multirow}
\usepackage{colortbl}
\usepackage[colorlinks,
            linkcolor=black,
            anchorcolor=blue,
            citecolor=blue
            ]{hyperref}
\newtheorem{thm}{Definition}

\begin{document}
	
\title{Fuzzy Information Entropy and Region Biased Matrix Factorization for Web Service QoS Prediction}

\author{
        \IEEEauthorblockN{
                Guoxing Tang\textsuperscript{}, 
                Yugen Du\textsuperscript{*}, 
                Xia Chen\textsuperscript{},
                Yingwei Luo\textsuperscript{},
                Benchi Ma\textsuperscript{}
            }
        \IEEEauthorblockA{
                \textsuperscript{}\textit{Software Engineering Institute, East China Normal University, Shanghai 200062, China}\\
                \textsuperscript{}\textit{Shanghai Key Laboratory of Trustworthy Computing} \\
                \textit{ygdu@sei.ecnu.edu.cn}
            }
        \thanks{DOI reference number: {10.18293/SEKE2024-023}}
    }


\IEEEpeerreviewmaketitle 
\maketitle
\IEEEpeerreviewmaketitle 
\begin{abstract}
Nowadays, there are many similar services available on the internet, making Quality of Service (QoS) a key concern for users. 
Since collecting QoS values for all services through user invocations is impractical, predicting QoS values is a more feasible approach. Matrix factorization is considered an effective prediction method. However, most existing matrix factorization algorithms focus on capturing global similarities between users and services, overlooking the local similarities between users and their similar neighbors, as well as the non-interactive effects between users and services. This paper proposes a matrix factorization approach based on user information entropy and region bias, which utilizes a similarity measurement method based on fuzzy information entropy to identify similar neighbors of users. Simultaneously, it integrates the region bias between each user and service linearly into matrix factorization to capture the non-interactive features between users and services. This method demonstrates improved predictive performance in more realistic and complex network environments. Additionally, numerous experiments are conducted on real-world QoS datasets. The experimental results show that the proposed method outperforms some of the state-of-the-art methods in the field at matrix densities ranging from 5\% to 20\%.
\end{abstract}


\section{Introduction}
With the continuous advancement and maturation of cloud computing technology, developers from enterprises and organizations are increasingly inclined to deploy their web services to the cloud, further driving the development of web services~\cite{duan2012survey,zheng2020web}. However, with the increasing number of web services on cloud platforms, there is a proliferation of homogeneous web services with similar or identical functionalities~\cite{xiong2018deep}. This presents a challenge for users in selecting services that suit their needs. In this scenario, QoS becomes a crucial consideration for users in selecting web services. Therefore, QoS-based service recommendation emerges as an important approach to address this issue, helping users better match their desired web services.

As one of the most popular model-based collaborative web service recommendation technology, matrix factorization (MF) is often used for QoS prediction~\cite{goldberg1992using,ghafouri2020survey,chen2022web}. It decomposes the user-service matrix into a user feature matrix and a service feature matrix, and the inner product of these two feature matrices is used to represent the prediction of the QoS values~\cite{luo2016large}. MF can capture the interactions between users and services and the global similarity between users and services well~\cite{zhong2022collaborative,lo2012collaborative,wu2022double,zhang2019unraveling,hsieh2017collaborative}. However, it struggles to capture the effect of non-interactions on the QoS values, such as resource capacity and service load, which heavily depend on service-specific factors and will cause prediction errors if not taken into account\cite{zhu2017online}, as well as ignores the local similarity between users and services, which is the similarity between a user or service and its similar neighbors\cite{zheng2020web}.

In this paper, we propose a fuzzy information entropy and region biased matrix factorization (FIEMF) which can capture the non-interactive features of users and services more fully by clustering users by region. Moreover, fuzzy information entropy (FIE) is used to measure the uncertainty of users' rating preferences, which globally makes full use of the users' rating information to find more similar neighboring users. Finally, it is demonstrated experimentally that the method proposed in this paper can achieve more effective QoS prediction.

In summary, this paper makes the following main contributions:
\begin{itemize}
    \item Users are clustered based on their regions to obtain more detailed bias centers. The location information of users and services is fully utilized to capture the effects of non-interactive features of users and services, making the model more consistent with real-world network scenarios.
    \item A FIE-based similarity measurement method is proposed. This method measures the uncertainty of user rating preferences using FIE, fully exploring the hidden rating preferences in user rating data. As a result, the discovered neighbors are more representative and better aligned with the rating habits of real-world users.
    \item Numerous experiments were conducted on real large-scale QoS dataset to evaluate FIEMF. The experimental results indicate that our proposed method outperforms the most advanced methods currently available.
\end{itemize}

The rest of the paper is organized as follows. Section II introduces the QoS prediction problem and the architecture of the methodology proposed in this paper. Section III describes in detail the methodology proposed in this paper. Section IV shows the experimental results. Section V describes the related work and Section VI concludes.

\section{Preliminaries}
\subsection{Problem Formulation}
Given a set of users $U$ and a set of services $S$, a user-service invocation record can be recorded as a triplet $\!<\!u,\!s, \!v\!>$, where $u  \in U$, $s \in S$, and $v$ is the QoS value obtained when user $i$ invokes service $j$. By collecting these invocation records, we can obtain a QoS matrix $Q$. Since in the real world most users only invoke a limited number of web services, the matrix $Q$ is sparse. The QoS prediction problem can be defined as: Given a sparse QoS matrix $Q$, predict the missing QoS values in the matrix $Q$.

As a commonly used method based on latent feature analysis (LFA), MF decomposes the QoS matrix $Q_{mn}$ into a user feature matrices $U_{mk}$ and a service feature matrices $S_{kn}$, and the inner product of these two matrices is used to predict the missing QoS values. $k$ represents the dimension of the feature. To determine the value of each element in the user feature matrix and service feature moments, we can minimize the following objective function:
\begin{equation}
\small{\begin{split}
         L =\min_{U,S} \frac{1}{2} \sum_{i=1}^{m} \sum_{j=1}^{n} I_{i j} (Q_{ij} - U_iS_j)^2 + \frac{\lambda}{2} \left( \|U\|_F^2 + \|S\|_F^2 \right),
\end{split}}
\end{equation}
where $I_{ij}$ returns 1 if there is a invocation record and 0 if there is not, $U_i$ and $S_j$ represent the feature vectors of user $i$ and service $j$, respectively.

Taking the effects of some factors specific to users and services into consideration, user bias and service bias are added to the model to improve the model performance. Therefore, the object function for bias-based MF can be derived as follows:
\begin{align}\label{eq:basic}
\begin{split}
         L =&\min_{U,S,b,p} \frac{1}{2} \sum_{i=1}^{m} \sum_{j=1}^{n} I_{i j} (Q_{ij} - U_iS_j - b_i - p_j)^2 \\
    &+ \frac{\lambda}{2}  \left( \|U\|_F^2 + \|S\|_F^2 + \|b\|_F^2 + \|p\|_F^2 \right),
\end{split}
\end{align}
where $b_i$ represents the bias specific to user $i$ and $p_j$ represents the bias specific to service $j$. $\| \cdot \|_F^2$ is \textit{Frobenius norm}\cite{recht2010guaranteed}. $\lambda$ is a parameter that controls the degree of regularization and is used to prevent overfitting.

\subsection{Method Review}
\begin{figure*}[h]
  \centering
  \includegraphics[width=1.0\textwidth]{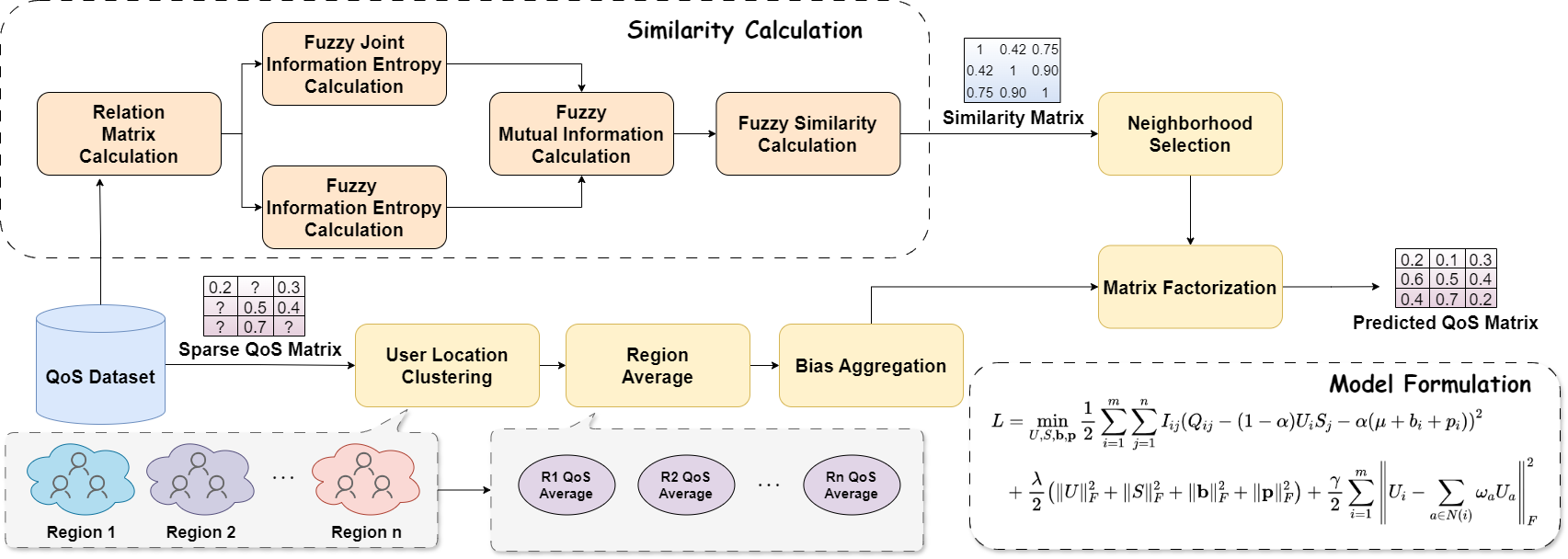}
  \caption{Overall framework of the FIEMF}
  \label{fig:FIEMF}
\end{figure*}

Traditional approaches cannot capture the hidden rating preferences of users in the QoS matrix, and the capture of bias is not sufficient. Therefore, FIEMF is proposed in this paper. The overall flow of FIEMF for integrating neighborhood information and bias information is shown in Fig.\textcolor{black}{~\ref{fig:FIEMF}}, which can be divided into 3 steps:
\begin{itemize}
    \item First, the users are clustered into regions according to their location information, then the QoS mean of each region is calculated as a bias center, and finally the bias of each user and service is integrated into the bias center.
    \item In order to get a measure of the users' rating preferences, the relationship matrix of each user is first calculated, then the FIE of each user and the fuzzy joint information entropy (FJIE) between users are derived from the relationship matrix. After that, the fuzzy mutual information (FMI) of the users is obtained base on FIE and FJIE, then the fuzzy similarity between the users is calculated based on FMI, and finally the neighbor selection is performed.
    \item Finally, we linearly combine the bias model into MF and incorporate a neighborhood regularization term based on FIE into the model to enhance the accuracy and robustness of the model.
\end{itemize}

The details of the process will be described in the next section.

\section{Proposed Approach}
\subsection{Neighborhood Selection}

User ratings of services can reflect their rating preferences. For instance, some users have stringent QoS requirements, leading to generally lower service ratings, while others tend to give higher ratings. Based on this, we combine all user rating information and propose a measurement method based on FIE to assess the uncertainty of user preferences.
\begin{thm}
Assuming U is the set of users, where user $u \in U$, and the set of service is denoted as $S = \{s_1,s_2,...,s_n \}$(n$\geq$2).Then the relationship matrix of user u on the service set S is
\begin{equation}\label{eq:3}
\boldsymbol{M}_{u}=\left(\begin{array}{cccc}
i_{11} & i_{12} & \cdots & i_{1 n} \\
i_{21} & i_{22} & \cdots & i_{2 n} \\
\vdots & \vdots & \ddots & \vdots \\
i_{n 1} & i_{n 2} & \cdots & i_{n n}
\end{array}\right),
\end{equation}
where $i_{xy} \in [0,1]$, representing the fuzzy equivalence relationship between service $s_x$ (1 $\leq$ x $\leq$ n) and service $s_y$ (1 $\leq$ y $\leq$ n) regarding user u. The calculation formula for $i_{xy}$ is defined as:
\begin{equation}
i_{xy} = \begin{cases}
exp(-\frac{1}{2} |r_{ux}-r_{uy}|)&,r_{ux}-r_{uy}<r_{med}\\
0&,otherwise,
\end{cases}
\end{equation}
where $r_{ux}$ is the rating of user u on service $s_x$ and $r_{med}$ is the median of the rating.
\end{thm}

In Eq.(\ref{eq:3}), we can gather information about their rating preferences by fully utilizing user ratings. If the difference between $s_x$ and $s_y$ is not less than the median rating $r_{med}$, it indicates that considering the ratings of user $u$ on services $s_x$ and $s_y$ separately does not help in determining the preferences of user $u$, $i_{xy}$ = 0. If the difference between $s_x$ and $s_y$ is less than $r_{med}$, we can infer that the ratings of services $s_x$ and $s_y$ to some extent reflect the rating preferences of user $u$. Specifically, if the ratings of user $u$ on services $s_x$ and $s_y$ are the same, then $i_{xy}$ = 1. 

After generating the relationship matrix $M_u$, we can calculate the FIE of user $u$ based on the fuzzy equivalence relations in the matrix $M_u$.

\begin{thm}
Given the relationship matrix of user u as ${M}_{u}$, The formula of fuzzy information entropy of user u is: 
\begin{equation}\label{eq:5}
\!F\!H(M_u) = -\frac{1}{n} \sum_{x=1}^{n} ln \frac{ \sum\limits_{y=1}^{n}i_{xy}}{n}.
\end{equation}
\end{thm}
FIE can reflect users' rating preferences. The closer the FIE between users, the more likely these two users have similar rating preferences, thereby indicating a higher similarity between them. Conversely, it is unlikely that these two users are neighbors to each other.

\begin{thm}
 If a and b are two users in the user set U, and the relationship matrices of $a$ and $b$ are represented by $M_a$ and $M_b$, then the fuzzy joint information entropy of a and b is obtained as:
 \begin{equation}\label{eq:6}
    \!F\!H(M_a,M_b) = -\frac{1}{n}\sum_{x=1}^{n}ln\frac{\sum\limits_{y=1}^{n}m_{xy}}{n},
 \end{equation}
 where $m_{xy} = min\{[i_{xy}]_a,[i_{xy}]_b\}$, representing the smaller value of the elements in the x-th row and y-th column of matrices $M_a$ and $M_b$.
\end{thm}

Expanding on this, we take the correlation between the rating preferences of pairwise users into consideration and propose a measurement method based on FMI to reflect the similarity between users.

\begin{thm}\label{fmi}
Based on Eq.(\ref{eq:5}) and Eq.(\ref{eq:6}), we can obtain the fuzzy mutual information between user $a$ and user $b$ to reflect the similarity of rating preferences between pairwise users. The formula is as follows:
\begin{equation}\label{eq:7}
    \!F\!M\!I(M_a,M_b) = \!F\!H(M_a) + \!F\!H(M_b) - \!F\!H(M_a,M_b).
\end{equation}
\end{thm}
If $a$ and $b$ have the same rating preferences, meaning $\!F\!H(M_a,M_b)$ = $\!F\!H(M_a)$ = $\!F\!H(M_b)$, then $\!F\!M\!I(M_a, M_b)$ achieves its maximum value. Conversely, if $a$ and $b$ have dissimilar rating preferences, $FMI(M_a, M_b)$ attains its minimum value.

To account for differences in FIE between users, we normalize the FMI and use the following formula instead of FMI for similarity calculations between users:
\small{
\begin{equation}\label{eq:8}
    sim_{\!F\!I\!E}(a,b) = \frac{2\!F\!M\!I(M_a,M_b)\!\times\!\exp{(-|FH(M_a)\!-\!F\!H(M_b)|)}}{\!F\!H(M_a)+\!F\!H(M_b)},
\end{equation}
}
where $sim_{\!F\!I\!E}(a,b)$ represents the FIE similarity of user $a$ and user $b$ with a value between [0,1], which is symmetric in nature.

After obtaining the similarity between the current user and other users, we select the Top-K similar users as neighbors. Therefore, for user $i$, its neighboring users can be identified by the following formula:
\begin{equation}\label{eq:9}
    N(i) = \{k|k\in Top\text{-}K(i),i\ne k\},
\end{equation}
where $Top\text{-}K(i)$ is the set of Top-K users with the highest similarity to the current user's FIE.

\subsection{Bias Integration}
Users in the same region tend to enjoy similar network facilities and communication environments, so their QoS values are more similar. Thus, each user in the same region has a smaller bias from the regional average, in order to capture this bias more accurately, we cluster the users by region, the region where user $i$ is located can be represented by the following equation:
\begin{equation}\label{eq:10}
    G(i) = \{x|x\in Region_i,x \ne i\},
\end{equation}
where $G(i)$ is the set of users in the same region as user $i$. When the region is too small, the QoS averages of the region are contingent and do not provide a good assessment of the local network conditions\cite{zheng2020web}, so we set the level of the region to the country level.

After clustering the users, we calculate the mean value of each user region. The mean value of the region of user $i$ can be calculated by:
\begin{equation}
    \mu(i) = \frac{\sum\limits_{u\in G(i)}R_i(u)}{|G(i)|},
\end{equation}
where $\mu(i)$ represents the QoS average value of the region of user $i$, $|G(i)|$ represents the number of services with rating records in the region, $R_i(u)$ represents the rating of each service by user $u$ in the region.

Considering the impact of user-specific and service-specific factors in the region, we add the user bias and service bias to $\mu(i)$ to obtain the following bias model:
\begin{equation}\label{eq:12}
    R_{ij} = \mu(i) +b_i +p_j,
\end{equation}
where $R_{ij}$ represents the QoS prediction of user $i$ invoking service $j$, $b_i,p_j$ is the bias of user $i$ and service $j$, respectively.

\subsection{Matrix Factorization}
Compared to Eq.(\ref{eq:basic}), FIEMF provides a more accurate delineation of the bias centers, followed by a linear integration of the bias information proposed in Eq.(\ref{eq:12}) into the MF to improve the accuracy and robustness of the predictions. Based on these facts discussed above, the prediction model proposed in this paper is:
\begin{equation}
    \hat{Q}_{ij} = \alpha U_iS_j + (1-\alpha)R_{ij},
\end{equation}
where $\hat{Q}_{ij}$ is the QoS prediction value of user $i$ invoking service $j$. Thus, the minimization objective function can be calculated by :
\begin{equation}\label{eq:14}
\begin{split}
         L =&\min_{U,S,b,p} \frac{1}{2} \sum_{i=1}^{m} \sum_{j=1}^{n} I_{i j} (Q_{ij} - \hat{Q}_{ij})^2\\ 
         &+ \frac{\lambda}{2}  \left( \|U\|_F^2 + \|S\|_F^2 + \|b\|_F^2 + \|p\|_F^2 \right),
\end{split}
\end{equation}
where $\lambda$ is the coefficient of the regularization term to prevent overfitting.

Eq.(\ref{eq:14}) can effectively estimate the overall structure associated with users or services (global similarity). However, the model performs poorly in detecting users or services that are similar to a small subset of users or services, and the neighborhood model can solve this problem well. Therefore, in this paper, a neighborhood regularization term based on FIE is added to Eq.(\ref{eq:14}), aiming at capturing the local similarity between users and services and reflecting the users' rating preferences. The objective function is as follows:
\begin{align}\label{eq:15}
\begin{split}
        L^\prime =& \min_{U,S,b,p} \frac{1}{2} \sum_{i=1}^{m}\sum_{j=1}^{n} I_{i j} (Q_{ij} - \hat{Q}_{ij})^2\\
        &+ \frac{\lambda}{2}  ( \|U\|_F^2 + \|S\|_F^2 + \|b\|_F^2 + \|p\|_F^2 )\\
        &+ \frac{\gamma}{2}\sum_{i=1}^m\|U_i -\sum\limits_{a\in N_i} \omega_aU_a\|_F^2,    
\end{split}
\end{align}

where $\gamma$ is the coefficient of the regular term, $N(i)$ is the Top-K similar users of user $i$, $\omega_a$ is the similarity weight of user $i$ to user $a$, which can be calculated by:
\begin{equation}
    \omega_a = \frac{sim_{\!F\!I\!E}(i,a)}{\sum\limits_{k \in N(i)} sim_{\!F\!I\!E}(i,k)},
\end{equation}
since the loss function is not convex, we use stochastic gradient descent (SDG)\cite{gemulla2011large} to solve for the optimal parameters of the $U,S,b,p$ with the following update rule:
\begin{align}
    U_i^ \prime =& U_i - \eta \frac{\partial L^\prime}{\partial U_i},\\
    S_j^ \prime =& S_i - \eta \frac{\partial L^\prime}{\partial S_j},\\
    b_i^ \prime =& b_i - \eta \frac{\partial L^\prime}{\partial b_i},\\
    p_i^ \prime =& p_j - \eta \frac{\partial L^\prime}{\partial p_j},
\end{align}

where $\eta$ is the learning rate of gradient descent, and the gradient of $U,S,b,p$ is calculated by:
\begin{align}
\begin{split}
        \frac{\partial L^\prime}{\partial U_i} =& \lambda U_i + \gamma(U_i -\sum\limits_{a\in N_i} \omega_aU_a) \\
        &- \alpha(Q_{ij} - \hat{Q}_{ij}) S_j,
\end{split}\\
        \frac{\partial L^\prime}{\partial S_j} =& \lambda S_j - \alpha(Q_{ij} - \hat{Q}_{ij}) S_j,\\
        \frac{\partial L^\prime}{\partial b_i} =& \lambda b_i - (1-\alpha)(Q_{ij} - \hat{Q}_{ij}),\\
        \frac{\partial L^\prime}{\partial p_j} =& \lambda p_j - (1-\alpha)(Q_{ij} - \hat{Q}_{ij}).
\end{align}

\section{Experiments}
In this section, we conduct several experiments on real QoS datasets to validate the effectiveness of FIEMF and verify its performance against 7 existing QoS prediction methods. The experiments have 3 main objectives: (1) Validating the validity and robustness of the FIEMF.
(2) Comparing the performance of FIEMF with other methods. (3) Analyzing the effect of parameters on performance. 

\subsection{Experimental environment}
We conducted experiments using Python3.7 and PyCharm on a computer with a 64-bit Windows 10 operating system using an AMD R7 6800H 3.2GHz CPU and 16GB RAM.

\subsection{Dataset}
We conducted experiments on the real-world QoS dataset WS-DREAM\footnote{https://github.com/wsdream}. It contains 1,974,675 QoS records from 339 distributed users from 30 different regions making invocations to 5,825 Web services in 73 different regions. The region information is included in the user list. We select response time as the evaluation metric for performance.

\subsection{Metrics}
The accuracy of QoS prediction directly reflects the effectiveness of the model. Therefore, we focuses on the accuracy of the model and uses mean absolute error (MAE) and root mean square error (RMSE) as evaluation metrics.
\begin{equation}
    MAE = \frac{1}{N}\sum_{i,j}|Q_{ij}-\hat{Q}_{ij}|,
\end{equation}
and RMSE is calculated as:
\begin{equation}
    RMSE = \sqrt{\frac{1}{N}\sum_{ij}(Q_{ij}-\hat{Q}_{ij})^2},
\end{equation}
where $Q_{ij}$ is the QoS value observed by user $i$ when user $i$ invokes service $j$ and $\hat{Q}_{ij})$ is the QoS prediction value, and $N$ represents the number of entries in the test set.

\subsection{Performance Comparison}
To validate the accuracy of our method, we compared FIEMF with the following 7 methods:
\begin{itemize}
    \item UMEAN: This method utilizes the known QoS values of the target user to find an average value.
    \item IMEAN: This method utilizes the known QoS values of the target service to find an average value.
    \item UIPCC: This method combines user-based CF and item-based CF.
    \item PMF: This method uses probabilistic matrix factorization to factorize QoS matrix for the prediction.
    \item BiasedMF: This method uses user bias and service bias.
    \item NIMF: This method integrates neighborhood information using Pearson Correlation Coefficient(PCC).
    \item NBMF: This method utilizes network bias to predict QoS.
\end{itemize}

In the real world, the user-service matrix is usually sparse, so we take 5\%, 10\%, 15\%, and 20\% of the QoS data from the original dataset as the training set and the rest as the test set. We set the parameters of the baseline method as a control experiment according to the corresponding paper. For our method, we set the parameters to $\alpha=0.15,\lambda=18,\gamma=18,d=10$. The maximum number of iterations of the model is set to 300.
\begin{table*}
\centering
\setlength{\extrarowheight}{2pt}
\caption{Acciracy Comparision}\label{table:1}
\begin{tabular}{c|ccccc|ccccc}
\hline
\multirow{2}{*}{Method}            & \multicolumn{4}{c}{MAE}                                       & \multirow{2}{*}{Improve} & \multicolumn{4}{c}{RMSE}                                      & \multicolumn{1}{l}{\multirow{2}{*}{Improve}} \\
                                   & D=5\%         & D=10\%        & D=15\%        & D=20\%        &                          & D=5\%         & D=10\%        & D=15\%        & D=20\%        & \multicolumn{1}{l}{}                         \\
\hline
UMEAN & 0.8816  & 0.8776  & 0.8743  & 0.8734  & 46.26\%   

& 1.8573    & 1.8558    & 1.8558    & 1.8579    & 32.57\% \\

IMEAN & 0.7036  & 0.6888  & 0.6848  & 0.6799  & 31.69\%      

& 1.5722    & 1.5382    & 1.5312    & 1.5297    & 18.91\% \\

UIPCC & 0.6398  & 0.5360  & 0.4876  & 0.4608  & 10.77\%      

& 1.4742    & 1.3461    & 1.2704    & 1.2216    & 5.77\% \\

PMF & 0.5686  & 0.4861  & 0.4512  & 0.4306  & 2.40\%      

& 1.5373    & 1.3143    & 1.2197    & 1.1695    & 4.16\% \\

BiasedMF & 0.5947  & 0.5124  & 0.4777  & 0.4559  & 7.44\%    

& 1.3822    & 1.2602    & 1.2086    & 1.1782    & 0.52\% \\

NIMF  & 0.5455  & 0.4817  & 0.4503  & 0.4287  & 1.02\%      

& 1.4659    & 1.2858    & 1.2088    & 1.1650    & 2.20\% \\

NBMF  & 0.5265  & 0.4827  & 0.4618  & 0.4488  & 1.94\%      

& 1.4255    & 1.2721    & 1.2235    & 1.1905    & 2.08\% \\

\hline
\textbf{FIEMF} & \textbf{0.5326} & \textbf{0.4752} & \textbf{0.4470} & \textbf{0.4302} & \textbf{-}            & \textbf{1.4079} & \textbf{1.2560} & \textbf{1.1893} & \textbf{1.1544} & \textbf{-} \\
\hline
\end{tabular}
\end{table*}

Table. \ref{table:1} shows the results of our experiments. Compared to all baseline methods, FIEMF improved MAE by 1.02\%$\sim$46.26\% and RMSE by 0.52\%$\sim$32.57\%. In terms of average improvement rates, our method outperforms all baseline methods in the range of 5\% to 20\% matrix density. Compared to BiasedMF, which takes into account user bias and service bias, MAE and RMSE are improved by 7.44\% and 0.52\%, respectively. However, when the density is 5\%, FIEMF performs slightly worse than BiasedMF in terms of RMSE. This is because the neighborhood regularization term in FIEMF can reduce the prediction performance of the bias term when the data is extremely sparse. Compared to the NBMF method that additionally considers network bias, the MAE and RMSE are improved by 1.94\% and 2.08\%, which is due to the fact that our method additionally utilizes the user's neighborhood information. Compared with NIMF, our method outperforms it in most cases, which indicates that our method is more reasonable for neighborhood selection. Moreover, we can also see that the accuracy of the prediction improves significantly as the density of the matrix increases, which indicates that the more QoS information is available, the more accurate the prediction is.

\subsection{Impact of Parameters}
In this part, we analyze the effect of different parameters on FIEMF, including the effect of dimension $d$, the effects of parameter $\gamma$ and parameter $\alpha$.

\begin{figure}
    \centering
    \includegraphics[width=0.24\textwidth]{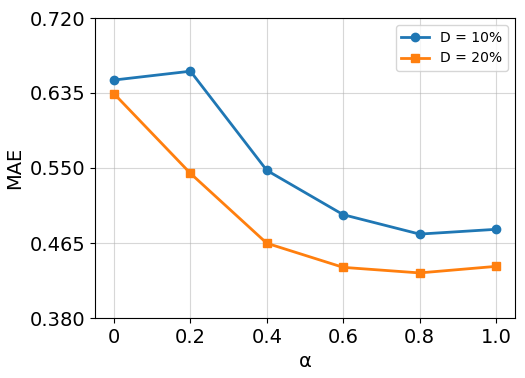}
    \includegraphics[width=0.24\textwidth]{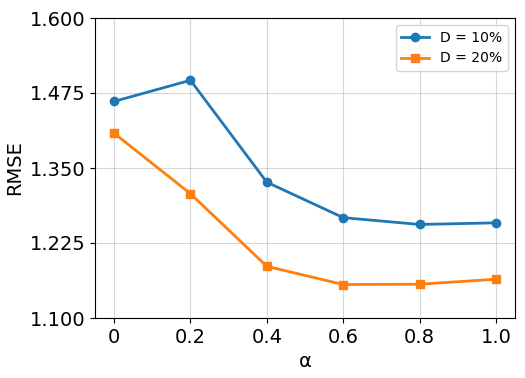}
    \caption{Impact of Parameter $\alpha$}\label{fig:2}
    \centering
    \includegraphics[width=0.24\textwidth]{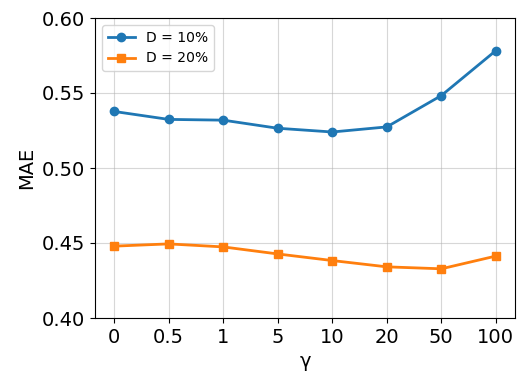}
    \includegraphics[width=0.24\textwidth]{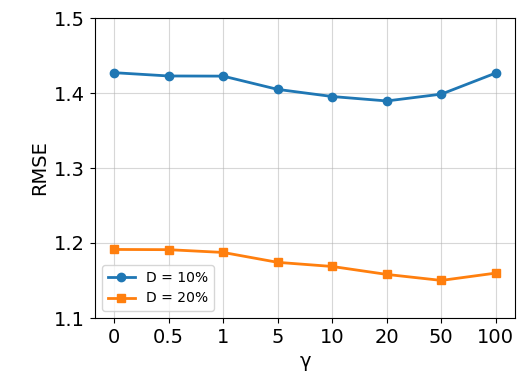}
    \caption{Impact of Parameter $\gamma$}\label{fig:3}
    \centering
    \includegraphics[width=0.24\textwidth]{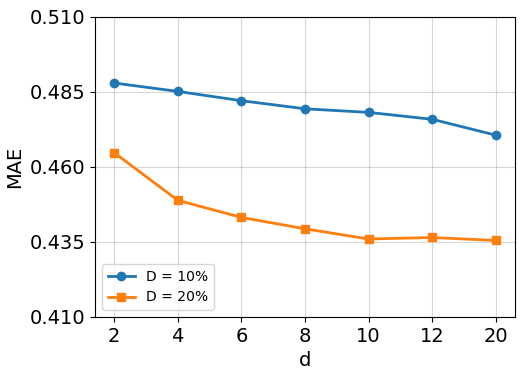}
    \includegraphics[width=0.24\textwidth]{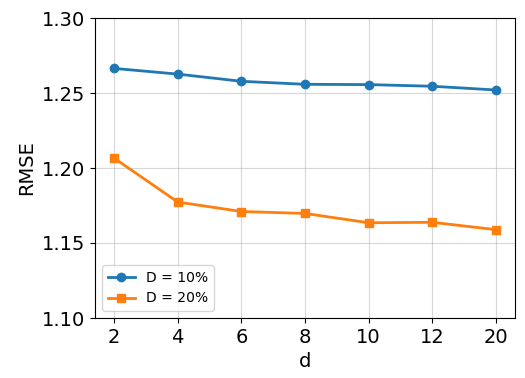}
    \caption{Impact of Parameter $d$}\label{fig:4}
\end{figure}

\begin{itemize}
    \item \textit{Impact of Parameter} $\alpha$: The parameter $\alpha$ controls the ratio of the weights of the user service interaction term to the region bias term. When $\alpha=0$, the prediction will completely rely on the region bias term without considering the effect of user service interaction. When $\alpha=1$, FIEMF degenerates into a MF with the addition of the neighborhood regular term. At this point, we can explore the improvement that the mere addition of the FIE-based neighborhood regular term brings to the traditional MF method. Fig.(\ref{fig:2}) shows the change in MAE and RMSE when $\alpha$ increases, the values of both MAE and RMSE are decreasing to a certain threshold and then increase, and when $\alpha=1$, the values of MAE and RMSE are much smaller than those of the PMF method under the corresponding density, which indicates that the inclusion of the FIE-based neighborhood regularity term significantly improves the accuracy of the prediction.
    \item \textit{Impact of Parameter} $\gamma$: The parameter $\gamma$ controls the weight of the FIE-based neighborhood regularization term. Fig.(\ref{fig:3}) shows the variation of MAE and RMSE when $\gamma$ varies from 0 to 100. It should be noted that when $\gamma=0$ , the FIEMF degenerates into a MF that merely incorporates the region bias, so we can explore whether the FIE-based neighborhood regular term improves the performance of the MF. It can be seen that when $\gamma$ increases, both MAE and RMSE first drop to a certain threshold and then start to rise, which indicates that the regular term improves the accuracy of prediction.
    \item \textit{Impact of Dimension} $d$: The dimension $d$ represents the number of latent features of the user service matrix, appropriate values of dimension can greatly improve the accuracy and efficiency of the prediction. To explore the impact of dimension, we vary the value of the dimension from 0 to 20. Fig.(\ref{fig:4}) demonstrates the changes in MAE and RMSE when the dimension varies. When the matrix density is 10\%, there is a slight decrease in MAE and RMSE, and when the matrix density is 20\%, both MAE and RMSE decrease, and the decrease slows down dramatically by $d = 10$. Since the computation of the user and service feature matrices increases dramatically when the dimension increases, we should balance accuracy and computational efficiency. When $d=10$, the efficiency and accuracy of the computation can be achieved with good results.
\end{itemize}

\section{ Related Work}
Collaborative filtering(CF) methods are often used for QoS prediction~\cite{zheng2020web}, including memory-based and model-based CF. Memory-based CF calculate the similarity between users using PCC~\cite{resnick1994grouplens} or cosine similarity, and then predict the QoS value of the target user based on the QoS values feedback from similar users. Model-based CF construct a global model using rating data, which has the ability to predict unknown data. 

Matrix factorization has become the most popular model-based CF technique due to its efficiency and scalability\cite{zhong2022collaborative,wu2022double}. Zhu et al\cite{zhu2017online}. proposed a MF method that takes into account the biases of users and services for capturing the effect of some user-specific and service-specific features, such as resource capacity and service load, on the QoS prediction values, but they ignored the fact that users in the same area often enjoy similar network facilities and communication environments, and also enjoy more similar bias centers. Tang et al.\cite{tang2019factorization} improved the classic factorization machine model by incorporating the location of service users. In order to fully utilize neighborhood information. Zheng et al.\cite{zheng2012collaborative} proposed a neighborhood integrated matrix factorization (NIMF), which tries to capture the local similarity of users and services by linearly combining the similar neighbor information of users into MF. However, the similarity measure they used only considers a small number of services where users have common rating items, so two users with large similarity do not necessarily have the same rating preferences. This leads to a high degree of chance in the prediction results~\cite{zhang2016entropy,jia2015target}. He et al.\cite{he2014location} used the K-means method to cluster based on the latitude and longitude information of user nodes and service nodes to optimize the neighborhood.

These methods have improved prediction accuracy to some extent, but they have not simultaneously considered the influence of user location information and non-interaction features, as well as user rating preferences.

\section{Conclusion}
In this paper, we propose a QoS prediction method FIEMF based on fuzzy information entropy and region bias. Firstly, clustering is performed based on the user's location information, and then bias information is added, followed by calculating the user's fuzzy information entropy, calculating the similarity according to the fuzzy information entropy, and then choosing the neighborhood based on the similarity, and finally, incorporating the bias model and the neighborhood model into MF in order to predict the QoS more accurately. The FIEMF model captures the non-interactive features of users and services more accurately by combining the user location information, and the fuzzy information entropy is utilized to measure the similarity between users, which can reflect the uncertainty of the user's rating preference well. Finally, this paper integrates neighborhood information into the MF model, which effectively exploits the local similarity in the MF model. In the future, we consider adding the time factor into our model to enable our model to perform real-time QoS prediction. In addition, we will consider better region partitioning, and more reasonable neighborhood selection to make our model more accurate and robust.

\bibliographystyle{elsarticle-num}
\bibliography{ref.bib}

\end{document}